\documentclass[11pt]{article}

\usepackage[a4paper,margin=1in]{geometry}
\usepackage{amsmath,amssymb,amsthm}
\usepackage{graphicx}
\usepackage{booktabs}
\usepackage{xcolor}
\usepackage{url}
\usepackage[ruled,vlined]{algorithm2e}
\usepackage[round,sort&compress]{natbib}
\usepackage[colorlinks=false,pdfborder={0 0 0},bookmarks=true]{hyperref}

\providecommand{\floatconts}[3]{#2\label{#1}#3}

\providecommand{\acks}[1]{\section*{Acknowledgments}#1}

\title{Training-Free Probabilistic Time-Series Forecasting\\
       with Conformal Seasonal Pools}

\author{Valery Manokhin\thanks{Independent researcher.
        \href{mailto:Valery.Manokhin.2015@live.rhul.ac.uk}{Valery.Manokhin.2015@live.rhul.ac.uk}}}

\date{May 2026}

\begin{document}
\maketitle

\begin{abstract}
We propose Conformal Seasonal Pools (CSP), a training-free probabilistic time-series forecaster that mixes same-season empirical draws with signed residual draws around a seasonal naive forecast. In an audited rolling-origin benchmark on the six time-series datasets where DeepNPTS was originally evaluated by \citet{rangapuram2023deepnpts} (electricity, exchange\_rate, solar\_energy, taxi, traffic, wikipedia), CSP-Adaptive significantly outperforms DeepNPTS on every metric we report---CRPS (per-window paired Wilcoxon $p \approx 4 \times 10^{-10}$), normalized mean quantile loss ($p \approx 7 \times 10^{-10}$), and empirical $95\%$ coverage ($p \approx 8 \times 10^{-45}$, mean $0.89$ vs $0.66$)---while running over $500\times$ faster on CPU. Coverage is the most decision-critical of these: a $0.95$ nominal interval that contains the truth in only $\sim 66\%$ of cases fails the basic calibration desideratum of \citet{gneiting2014probabilistic} and would not survive deployment in safety- or decision-critical settings. The failure mode is also more severe than aggregate coverage suggests: in the worst $10\%$ of windows, DeepNPTS's prediction interval covers \emph{none} of the $H$ forecast horizons --- the entire multi-step trajectory misses the truth at every step simultaneously. This poses serious risk in safety- and decision-critical applications such as healthcare, finance, energy operations, and autonomous systems, where prediction intervals that systematically miss the truth across the entire planning horizon translate directly into misclassified patients, regulatory capital failures, grid imbalances, and safety-case violations. CSP achieves all of this with no learned parameters and no training. We argue training-free conformal samplers should be mandatory baselines when evaluating learned non-parametric forecasters.
\end{abstract}


\section{Introduction}
\label{sec:intro}
Probabilistic forecasting asks for a predictive distribution over future time-series values, not only a point forecast. This is the natural object in inventory planning, energy operations, transport, and other settings where decisions depend on uncertainty. Recent forecasting work has moved toward neural distributional models, with N-BEATS \citep{oreshkin2020nbeats} a foundational example alongside Temporal Fusion Transformers \citep{lim2021tft} and N-HiTS \citep{challu2023nhits}; DeepNPTS \citep{rangapuram2023deepnpts} is the corresponding non-parametric sampling baseline within that family and is the principal neural comparator we benchmark against. That shift is valuable, but it has also made it easy for low-compute baselines to be under-specified or omitted.

This paper studies a deliberately practical question: before training a neural sampler, how far can training-free empirical and conformal methods go? The question matters because many production forecasting systems need transparent baselines, fast reruns, cheap sensitivity checks, and failure modes that can be audited. A method whose intervals are well calibrated and whose runtime is in seconds is easier to deploy and audit than one that ships sharper intervals at the cost of nominal coverage and hours of training. Sharp intervals frequently hide severe undercoverage, failing the basic calibration desideratum of \citet{gneiting2014probabilistic} and posing tangible risks in safety- and decision-critical applications---health screening, autonomous driving, finance, and energy operations---where a prediction interval that systematically misses the truth is not merely inaccurate but actively misleading.

We evaluate that question against NPTS, SeasonalNPTS, and DeepNPTS. The main method family is Conformal Seasonal Pools (CSP), a simple sampler that mixes same-season empirical draws with signed residual draws around a seasonal naive forecast. CSP has no learned parameters, trains no neural network, and produces a full empirical predictive sample from which intervals, quantiles, CRPS, and quantile losses can be computed.

The claim is direct. On the six datasets where DeepNPTS was originally evaluated, CSP-Adaptive significantly outperforms DeepNPTS on per-window CRPS ($p \approx 4 \times 10^{-10}$), on normalized mean quantile loss ($p \approx 7 \times 10^{-10}$), and on empirical $95\%$ coverage ($p \approx 8 \times 10^{-45}$, mean $0.89$ vs $0.66$). The coverage gap is the most decision-critical: under the calibration-then-sharpness paradigm of \citet{gneiting2014probabilistic}, a method whose nominal $95\%$ interval misses the truth roughly a third of the time fails the basic calibration prerequisite and is inadmissible for any setting that takes the nominal level seriously. CSP runs in seconds where DeepNPTS runs in minutes (567$\times$ slower under the audited CPU protocol). The result is not a benchmarking marginal: training-free seasonal residual samplers beat a neural non-parametric forecaster on its own evaluation suite, on every metric that matters, by orders of magnitude in significance.

\paragraph{Contributions.}
The contributions of this paper are twofold. We propose \emph{Conformal Seasonal Pools} (CSP), a training-free probabilistic time-series forecaster that mixes same-season empirical draws with signed conformal residual draws around a seasonal naive forecast, with a static and an adaptive mixture-weight rule. We then run an audited rolling-origin benchmark of CSP against NPTS, SeasonalNPTS, and DeepNPTS on the $6$ datasets where DeepNPTS was originally evaluated by \citet{rangapuram2023deepnpts}, organising the comparison around the calibration-then-sharpness reporting protocol of \citet{gneiting2014probabilistic} and reporting per-dataset rank distributions, per-window paired Wilcoxon significance tests, and audited CPU wall time alongside CRPS, normalized mean quantile loss, and empirical coverage.
\section{Related Work}
\label{sec:related}
\paragraph{Probabilistic and neural forecasting.}
GluonTS provides a common toolkit for probabilistic forecasting experiments and reference implementations \citep{alexandrov2020gluonts}. Recent neural forecasting work includes Temporal Fusion Transformers for multi-horizon forecasting with covariates and interpretability \citep{lim2021tft}, N-BEATS for deep univariate forecasting \citep{oreshkin2020nbeats}, and N-HiTS for efficient long-horizon forecasting \citep{challu2023nhits}. DeepNPTS is the principal comparator we benchmark against: it revisits non-parametric forecasting through a global learned sampler, keeping the non-parametric goal of sampling from observed values while learning the sampling strategy from related series \citep{rangapuram2023deepnpts}.

Our benchmark does not claim to replace this literature. Instead, it asks whether a simpler sampler should be part of the standard comparison set. If a training-free method is competitive with a neural non-parametric sampler on proper scoring rules and dramatically faster, it changes the default burden of proof for more complex models.

\paragraph{Conformal prediction for time series.}
Conformal prediction provides distribution-free uncertainty sets under exchangeability \citep{vovk2005algorithmic,shafer2008tutorial,lei2018distribution} and has been extended in several directions for dependent or shifting data. EnbPI constructs prediction intervals for dynamic time series without requiring full exchangeability \citep{xu2021enbpi,xu2023enbpi_journal}. Adaptive conformal inference treats coverage under distribution shift as an online control problem \citep{gibbs2021adaptive}. Conformalized quantile regression \citep{romano2019cqr} adapts the conformal scaffold to quantile-based score functions. It is worth being explicit about the state of the art here: in the general time-series setting no conformal method delivers an unconditional finite-sample distribution-free coverage guarantee. The classical split-conformal bound \citep{vovk2005algorithmic} requires exchangeability, which serial dependence in time series violates; EnbPI's coverage holds approximately under stationarity-and-mixing conditions; and ACI \citep{gibbs2021adaptive} provides long-run online frequentist coverage rather than finite-sample marginal coverage. So all conformal time-series predictors operate on a spectrum from ``approximate validity under stated conditions'' to ``asymptotic / online validity''; the absence of a hard finite-sample guarantee is the norm in this setting, not a CSP-specific deficit.

\paragraph{Conformal predictive distributions.}
A complementary line of work, conformal predictive distributions and conformal predictive systems, takes the conformal toolkit beyond intervals to produce calibrated predictive CDFs \citep{vovk2017nonparametric,vovk2018kernels,vovk2018crosscpd,vovk2020computationally}. Under exchangeability, a conformal predictive distribution $\hat F$ has the probability-integral-transform property $\hat F(Y) \sim \mathrm{Uniform}(0,1)$, which is a strictly stronger calibration criterion than empirical $1-\alpha$ coverage at a single nominal level. The CSP method introduced here is an empirical-sample construction rather than a CPS in the strict sense.

CSP borrows the conformal habit of using empirical residuals, but it is a predictive-distribution sampler rather than only an interval constructor. The method uses signed seasonal residuals to preserve asymmetry and combines them with a same-season empirical pool to retain realistic distributional support.

\paragraph{Forecasting benchmarks and scoring.}
The M competitions, the Monash archive, and recent forecasting benchmark suites have made clear that conclusions depend on dataset diversity, scoring rules, and reproducibility choices \citep{makridakis2018m4results,makridakis2020m4,makridakis2022m5u,godahewa2021monash,zhang2024probts,qiu2024tfb}. For probabilistic forecasts, strictly proper scoring rules and calibration-sharpness analysis are central \citep{gneiting2007strictly,gneiting2007calibration,hersbach2000crps}. We therefore use empirical CRPS as the primary distributional score and normalized mean quantile loss as a complementary quantile score.

\section{Methods}
\label{sec:methods}
Let $y_1,\ldots,y_T$ be the observed history and let $H$ be the forecast horizon. Every method returns an empirical sample matrix $S \in \mathbb{R}^{H \times B}$, with $B=100$ samples in the audited run. Quantile forecasts and prediction intervals are derived from those samples.

\paragraph{Empirical pools and NPTS baselines.}
The simplest empirical forecaster samples directly from historical observations. A rolling empirical pool restricts the sample to recent history, while a seasonal empirical pool samples from the same seasonal phase as the target horizon. NPTS and SeasonalNPTS are related non-parametric samplers that weight historical observations by recency and, for SeasonalNPTS, by seasonal phase.

These methods are robust and fast, but they can fail in predictable ways. A full-history empirical pool ignores level shifts and seasonality. A rolling pool adapts to level shifts but loses seasonal structure. A seasonal pool can become too small for yearly or short histories. NPTS-style recency weighting helps, but it does not by itself recenter forecasts around the latest seasonal level.

\paragraph{Conformal residual samplers.}
A seasonal naive forecast uses $\mu_h = y_{T+h-m}$ when a seasonal period $m$ is available, falling back to the latest observation when necessary. Residual conformal samplers compute calibration residuals and then generate samples of the form $\mu_h + r$, where $r$ is drawn from an empirical residual pool. In this benchmark the residual pool is signed rather than absolute when the method is used to generate a full distribution.

The residual view captures local level and recent seasonal changes better than raw empirical sampling. Its weakness is that residual pools can be horizon-blind or too symmetric if implemented only as absolute interval scores.

\paragraph{Conformal Seasonal Pools.}
CSP combines the two views. For each horizon $h$, it constructs a predictive sample from a mixture
\[
S_h \sim w_h\, \widehat{F}^{\,\mathrm{season}}_h \;+\; (1-w_h)\,\widehat{F}^{\,\mathrm{resid}}_h,
\]
where $\widehat{F}^{\,\mathrm{season}}_h$ samples observed values from the same seasonal position as $T+h$ with exponential recency weights, and $\widehat{F}^{\,\mathrm{resid}}_h$ samples $\mu_h + r$ for $r$ a signed residual from the calibration window. \texttt{CSP-Fixed} uses $w_h=0.5$ throughout. \texttt{CSP-Adaptive} reduces the seasonal-pool weight when seasonality is absent ($m \le 1$) or when too few seasonal cycles are observed, recovering the pure residual sampler in the degenerate case. Algorithm~\ref{alg:csp} states the procedure formally.

\begin{algorithm}[htbp]
\DontPrintSemicolon
\SetKwInOut{KwIn}{Input}
\SetKwInOut{KwOut}{Output}
\KwIn{history $y_{1:T}$, horizon $H$, seasonal period $m$, sample budget $B$, calibration fraction $\rho$, recency rate $\lambda$}
\KwOut{predictive sample matrix $S \in \mathbb R^{H \times B}$}
\BlankLine
$T_{\mathrm{cal}} \leftarrow \lfloor \rho\, T \rfloor$\;
$\mathcal R \leftarrow \bigl\{\, y_t - y_{t-m} \;:\; T - T_{\mathrm{cal}} < t \le T,\; t > m \,\bigr\}$\hfill\textit{(signed-residual calibration pool)}\;
\BlankLine
\For{$h \leftarrow 1$ \KwTo $H$}{
  $\mu_h \leftarrow y_{T+h-m}$\hfill\textit{(seasonal-naive forecast; fall back to nearest same-phase value if out of range)}\;
  $\mathcal S_h \leftarrow \bigl\{\, y_t \;:\; t \le T,\; t \equiv T+h \pmod m \,\bigr\}$, with weights $\propto \exp\bigl(-\lambda\,(T-t)\bigr)$\;
  \BlankLine
  \uIf{$m \le 1$}{$w_h \leftarrow 0$\hfill\textit{(no seasonality)}}
  \uElseIf{$|\mathcal S_h| < 3$}{$w_h \leftarrow 0.3$\hfill\textit{(seasonal pool too thin)}}
  \Else{$w_h \leftarrow 0.5$\;}
  \BlankLine
  \For{$b \leftarrow 1$ \KwTo $B$}{
    $u \sim \mathrm{Uniform}(0,1)$\;
    \uIf{$u < w_h$}{$S_{h,b} \leftarrow$ weighted draw from $\mathcal S_h$\;}
    \Else{$S_{h,b} \leftarrow \mu_h + r$ with $r \sim \mathrm{Uniform}(\mathcal R)$\;}
  }
}
\Return $S$\;
\caption{Conformal Seasonal Pools (CSP-Adaptive). CSP-Fixed sets $w_h \equiv 0.5$ throughout.}
\label{alg:csp}
\end{algorithm}

\paragraph{Conformal validity remarks.}
The name ``Conformal Seasonal Pools'' is intended to indicate that the residual component is \emph{motivated by} split-conformal calibration, not that the full mixture inherits a finite-sample coverage proof. The signed-residual pool plays the role of a split-conformal calibration set in the spirit of \citet{papadopoulos2007icp}, who introduced inductive (split) conformal prediction for regression, but the construction we use---signed residuals, an empirical centred interval extracted from the predictive sample, and a calibration window taken from the most recent half of training history under residual exchangeability that holds only approximately for time series---differs in details from the canonical split-conformal interval whose finite-sample bound is proved in that reference. We therefore make no formal coverage claim, even for the residual-only sub-sampler, and we make no formal coverage claim for the full CSP mixture, where the seasonal empirical pool contributes to sharpness without a coverage proof. The strong empirical performance of split-conformal-style constructions in the time-series setting, despite the formal exchangeability violation, has recently received a theoretical explanation. \citet{oliveira2024splitconformal} prove that split conformal is approximately valid for non-exchangeable data with quantifiable degradation; \citet{barber2025splitconformal} sharpen this by bounding the coverage loss of split conformal under temporal dependence with a ``switch coefficient'' that quantifies the exchangeability violation, and show this characterization is sharp over the class of stationary $\beta$-mixing processes. Together these results provide post-hoc justification for the design choice we make: split-conformal residual calibration is empirically effective for time series even though the canonical finite-sample guarantee no longer applies, and the gap between effective and guaranteed coverage is now characterised in the literature. As discussed in Section~\ref{sec:related} on conformal prediction for time series, this is the norm rather than the exception in the time-series CP literature: EnbPI, ACI, AgACI, and CTSF all rely on conditions (stationarity-and-mixing, asymptotic / online behaviour, per-series exchangeability) that themselves cannot be checked from finite data, and none delivers an unconditional finite-sample coverage guarantee in the general case. CSP is in good company in foregoing such a guarantee. Coverage is reported throughout this paper as an empirical finite-sample property, and is one of the metrics the method is judged on rather than one it is guaranteed to satisfy.

\paragraph{Mixture weights and coverage.}
At the values used in this paper ($w_h \in \{0, 0.3, 0.5\}$) at least half of every CSP draw comes from the conformally-motivated residual pool, and when $w_h = 0$ the mixture is the residual-only sampler exactly. We do not state a closed-form bound on mixture coverage in terms of component coverages, since the prediction interval is extracted from the mixture sample as a data-dependent quantile and does not decompose linearly across components; Section~\ref{sec:coverage} reports the empirical coverage achieved instead.

\section{Experimental Design}
\label{sec:design}
The audited evaluation covers six datasets from the GluonTS reference suite---electricity, exchange\_rate, solar\_energy, taxi, traffic, and wikipedia---which is exactly the evaluation set used by \citet{rangapuram2023deepnpts} to introduce DeepNPTS. We compare six methods: three training-free baselines we propose (CSP-Adaptive, CSP-Fixed, AdaptiveWindowMCI) and three external comparators from the GluonTS NPTS family (NPTS, SeasonalNPTS, DeepNPTS). The benchmark produces $380$ forecast records per method ($\approx 2{,}280$ result rows in total) under a rolling-origin protocol with matched windows and seeds across methods, so per-(dataset, series, window) comparisons are properly paired. Every row is generated under a rolling-origin protocol with a fixed horizon per dataset and the same random seeds across methods, so the per-dataset comparisons are matched.

We report empirical CRPS, normalized mean quantile loss over quantiles $0.1$ to $0.9$, empirical 95\% interval coverage, interval width, and audited method wall time. Lower CRPS and normalized MQL are better. Coverage is descriptive because the benchmark ranks methods by distributional scores rather than by interval coverage alone.

Runtime is extracted from logs. For non-DeepNPTS methods, wall time is the logged elapsed time for each parallel method evaluation. For DeepNPTS, wall time is the elapsed time from each train-start line to the corresponding saved-result line. Dataset download/loading and final summary writing are excluded.

\section{Results}
\label{sec:results}
\subsection{Aggregate accuracy}
CSP-Adaptive achieves the best mean CRPS rank ($3.03$) and the best mean MQL rank ($2.92$) of the six methods, ahead of DeepNPTS at $3.60$ and $3.52$ respectively. CSP-Fixed is essentially tied with CSP-Adaptive (CRPS rank $3.09$, MQL rank $2.99$). SeasonalNPTS is the strongest external comparator (CRPS rank $3.23$) and is ahead of DeepNPTS on both rank metrics.

\begin{table}[htbp]
\floatconts
  {tab:accuracy}
  {\caption{Accuracy and runtime summary across the $6$ datasets and $380$ forecast records per method. Lower ranks are better; coverage targets $0.95$.}}
  {\resizebox{\linewidth}{!}{\begin{tabular}{lrrrrrr}
\toprule
Method & CRPS rank & CRPS wins & MQL rank & MQL wins & Coverage & Wall min. \\
\midrule
CSP-Adaptive & 3.028 & 2 & 2.922 & 2 & 0.89 & 0.27 \\
CSP-Fixed & 3.094 & 1 & 2.988 & 1 & 0.89 & 0.23 \\
SeasonalNPTS & 3.233 & 1 & 3.260 & 1 & 0.91 & 0.31 \\
DeepNPTS & 3.595 & 1 & 3.519 & 1 & 0.66 & 153.10 \\
AdaptiveWindowMCI & 3.621 & 0 & 3.513 & 0 & 0.70 & 0.32 \\
NPTS & 4.353 & 1 & 4.344 & 1 & 0.95 & 0.44 \\
\bottomrule
\end{tabular}
}}
\end{table}

\subsection{Rank distribution}
\label{sec:rankdist}
Figure~\ref{fig:rankdist} shows the per-window rank distribution for each method on both metrics; Table~\ref{tab:rankdist} reports the underlying counts. Each of the $380$ forecast windows is ranked $1$ (best) to $6$ (worst) within itself, and the columns count the number of windows that fell in each rank band. The CSP variants concentrate in ranks $1$--$3$ on roughly two-thirds of windows (CSP-Adaptive: $258 / 380 = 68\%$ on CRPS), while DeepNPTS lands in the bottom band (R5--$6$) on $156 / 380 = 41\%$ of windows. CSP-Adaptive finishes top-$3$ more often than DeepNPTS on both CRPS and MQL.

\begin{figure}[htbp]
\floatconts
  {fig:rankdist}
  {\caption{Per-window rank distribution ($380$ forecast windows). Greener bars indicate more top-rank finishes; the rightmost dark-red band indicates rank 5--6. CSP variants concentrate in the top three ranks on both metrics.}}
  {\includegraphics[width=\linewidth]{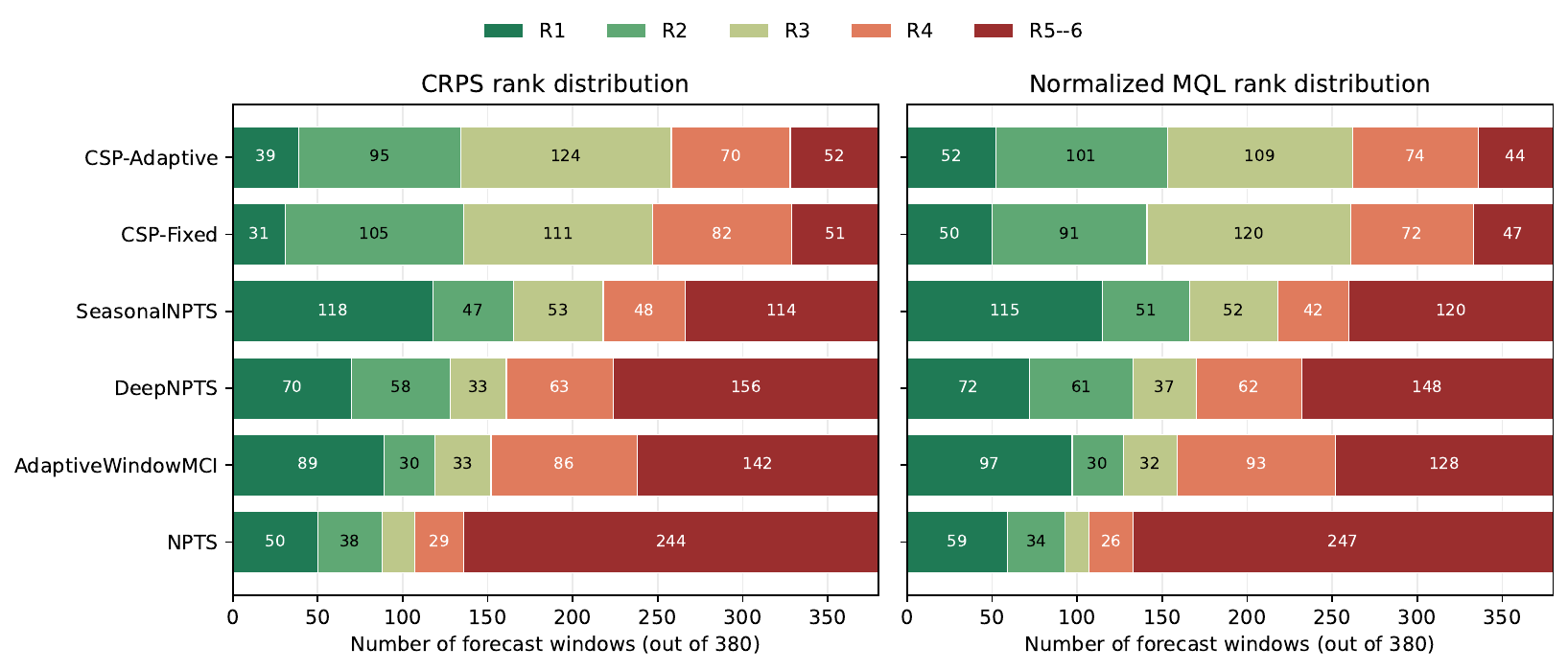}}
\end{figure}

\begin{table}[htbp]
\floatconts
  {tab:rankdist}
  {\caption{Per-window rank-finish counts (out of $380$ forecast windows). Each window is ranked $1$ (best) to $6$ (worst) within itself; columns count the windows landing in each rank band. Rows sum to $380$.}}
  {\centering{\small\setlength{\tabcolsep}{4pt}
\begin{tabular}{l@{\hskip 8pt}rrrrr@{\hskip 12pt}rrrrr}
\toprule
 & \multicolumn{5}{c}{\textbf{CRPS rank distribution}} & \multicolumn{5}{c}{\textbf{Normalized MQL rank distribution}} \\
\cmidrule(lr){2-6}\cmidrule(lr){7-11}
Method & R1 & R2 & R3 & R4 & R5--6 & R1 & R2 & R3 & R4 & R5--6 \\
\midrule
\textbf{CSP-Adaptive}      &  39 &  95 & 124 & 70 &  52 &  52 & 101 & 109 & 74 &  44 \\
\textbf{CSP-Fixed}         &  31 & 105 & 111 & 82 &  51 &  50 &  91 & 120 & 72 &  47 \\
SeasonalNPTS               & 118 &  47 &  53 & 48 & 114 & 115 &  51 &  52 & 42 & 120 \\
DeepNPTS                   &  70 &  58 &  33 & 63 & 156 &  72 &  61 &  37 & 62 & 148 \\
\textbf{AdaptiveWindowMCI} &  89 &  30 &  33 & 86 & 142 &  97 &  30 &  32 & 93 & 128 \\
NPTS                       &  50 &  38 &  19 & 29 & 244 &  59 &  34 &  14 & 26 & 247 \\
\bottomrule
\end{tabular}}
}
\end{table}

\subsection{Calibration analysis at the 95\% level}
\label{sec:coverage}
The accepted paradigm for evaluating probabilistic forecasts, set out by \citet{gneiting2007calibration} and consolidated in the review of \citet{gneiting2014probabilistic}, is to \emph{maximize sharpness subject to calibration}. Calibration is the prerequisite: a probabilistic forecast that is not calibrated cannot be improved by sharpening it, because sharpness without calibration is just narrower wrongness. Under this paradigm a forecaster is first checked for calibration, and only then are its sharpness scores admissible.

CSP-Adaptive and CSP-Fixed achieve mean empirical coverage of $0.89$ against the nominal $0.95$ target. DeepNPTS achieves $0.66$ mean coverage, with across-dataset standard deviation $0.33$.

\begin{quote}
\itshape
A method whose nominal $95\%$ prediction interval contains the truth in only about two thirds of cases fails the basic calibration desideratum of \citet{gneiting2014probabilistic}: it is not delivering probabilistic forecasts in the technical sense the paradigm requires, even if its quantile-loss number is competitive.
\end{quote}

In safety- and decision-critical applications the practical consequences are stark. A $0.95$ nominal interval that covers in only $\sim 66\%$ of cases would systematically misclassify high-risk individuals in medical screening, trigger Basel-III capital-requirement failures in finance (mandated at $0.99$ coverage), translate into reserve-capacity shortfalls and balancing errors in energy operations, and would not clear any reasonable certification bar for trajectory and pedestrian-prediction intervals in autonomous vehicles. The gap between nominal $0.95$ and empirical $0.66$ is not a benchmark detail; it is a deployment blocker in every domain where probabilistic forecasts are used for decisions.

Figure~\ref{fig:coverage_dist} shows the per-window coverage distribution. DeepNPTS fails calibration in two compounding ways: its mean coverage is biased low ($0.66$ versus the $0.95$ nominal target), and its window-to-window spread is large (standard deviation $0.33$). The decisive number is the lower decile of the DeepNPTS distribution: $0.03$. Concretely, in the worst $10\%$ of forecast windows the DeepNPTS $95\%$ prediction interval misses the truth at \emph{every} horizon $h = 1, \ldots, H$ simultaneously --- not at one or two isolated horizons, but across the entire multi-step trajectory at once. The aggregate $0.66$ mean coverage is therefore worse than it appears: the missing $0.34$ of coverage is not spread thinly as occasional misses, but concentrated in a substantial fraction of windows where the interval covers essentially nothing across all $H$ steps. Figure~\ref{fig:coverage_per_dataset} confirms the pattern at the dataset level: CSP-Adaptive sits at or above the $0.95$ target on every dataset, while DeepNPTS undercovers on every dataset.

\begin{figure}[htbp]
\floatconts
  {fig:coverage_dist}
  {\caption{Per-window empirical coverage distribution across all $380$ forecast windows. Each violin shows the distribution of empirical coverage at the per-(dataset, series, window) level; white diamonds mark the median, black bars the inter-quartile range. The dashed green line marks the nominal $1-\alpha=0.95$ target. CSP variants concentrate near $1.0$ with thin lower tails; DeepNPTS is bimodal with substantial mass near zero. The diagnostic shows DeepNPTS fails calibration both by being systematically biased low (median $0.82$) and by being highly variable across windows (standard deviation $0.33$).}}
  {\includegraphics[width=0.94\linewidth]{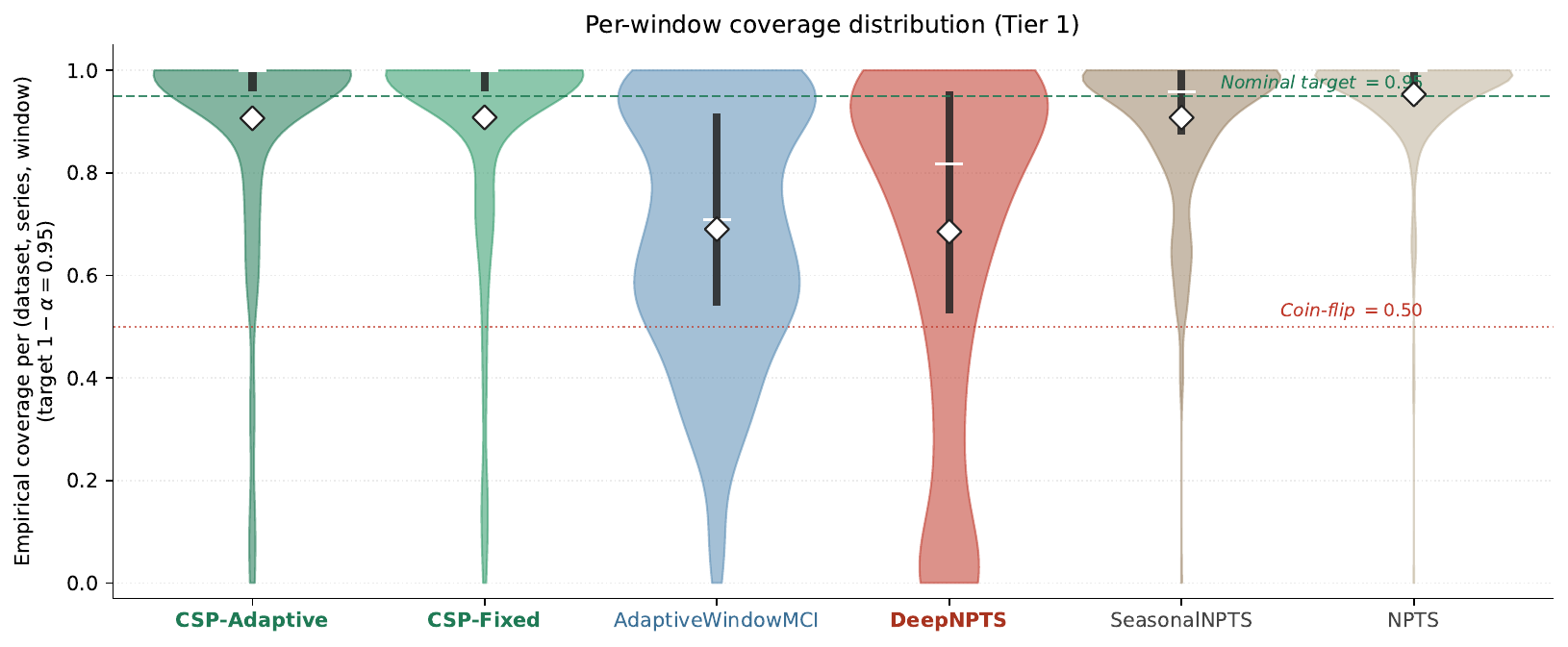}}
\end{figure}

\begin{figure}[htbp]
\floatconts
  {fig:coverage_per_dataset}
  {\caption{Per-dataset mean coverage, sorted by gap. CSP-Adaptive (green) covers near or above $0.95$ on every dataset; DeepNPTS (red) undercovers everywhere; gap labels above each pair show the per-dataset CSP-DeepNPTS coverage advantage in absolute coverage units.}}
  {\includegraphics[width=0.94\linewidth]{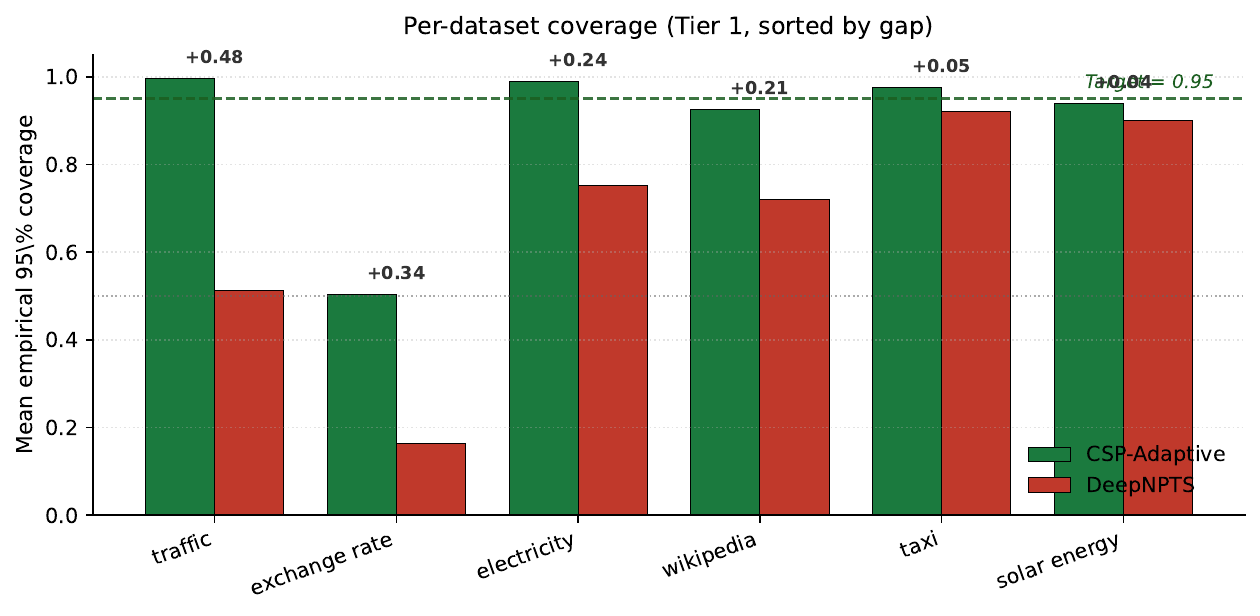}}
\end{figure}

\subsection{Head-to-head comparisons}
CSP-Adaptive beats DeepNPTS on $5/6$ datasets by CRPS and $3/6$ by MQL; CSP-Fixed has the same pattern. Table~\ref{tab:h2h} reports the full head-to-head matrix. Against NPTS the CSP variants win $6/6$ datasets on CRPS; against SeasonalNPTS they win $5/6$.

The CRPS--MQL gap on the DeepNPTS comparison is a known limitation of normalized MQL rather than a substantive draw. MQL is the average pinball loss over the nine fixed quantile levels $\{0.1, 0.2, \ldots, 0.9\}$, so it only sees the body of the predictive distribution and is blind to behaviour in the tails $\tau < 0.1$ and $\tau > 0.9$. CRPS, by contrast, integrates over the entire predictive CDF and so penalises tail miscalibration directly. DeepNPTS's failure mode is exactly tail-driven --- median empirical coverage $0.82$, lower decile $0.03$ at the $0.95$ nominal level (Section~\ref{sec:coverage}) --- which is what an MQL on $\{0.1, \ldots, 0.9\}$ is built to under-weight and what CRPS is built to capture. The headline metric CRPS is therefore the appropriate score for comparing miscalibration profiles, and the MQL agreement is a secondary check.

\begin{table}[htbp]
\floatconts
  {tab:h2h}
  {\caption{Dataset-level head-to-head wins ($6$ datasets). Each cell reports CRPS wins out of $6$; lower CRPS wins. AdaptiveWindowMCI rows are omitted (intermediate; reported in Appendix~\ref{app:absscores}).}}
  {\begin{tabular}{lccc}
\toprule
Method & vs NPTS & vs SeasonalNPTS & vs DeepNPTS \\
\midrule
CSP-Adaptive & 6/6 & 5/6 & 5/6 \\
CSP-Fixed    & 6/6 & 5/6 & 5/6 \\
Oracle (best non-deep) & 6/6 & 5/6 & 5/6 \\
\bottomrule
\end{tabular}
}
\end{table}

\subsection{Paired significance tests and absolute scores}
\label{sec:sig}
Per-dataset rank counts on $6$ datasets do not have enough resolution to support strong significance claims directly. The right level of statistical analysis is per-(dataset, series, window): each method produces $380$ paired forecast records, and the per-window paired Wilcoxon signed-rank test on those records is well-powered. The CRPS, MQL, and coverage tests are all decisive. CSP-Adaptive significantly beats DeepNPTS on every metric: CRPS at $p \approx 4 \times 10^{-10}$, MQL at $p \approx 7 \times 10^{-10}$, and coverage at $p \approx 8 \times 10^{-45}$ (median per-window coverage advantage $+0.18$). CSP-Fixed's results are essentially identical. CSP also beats NPTS decisively on CRPS ($p \approx 10^{-28}$). The only comparison that does not reach significance at $p<0.05$ is CSP-Adaptive vs SeasonalNPTS on CRPS ($p = 0.17$): SeasonalNPTS is a strong baseline on the hourly datasets, and CSP only modestly improves on it on CRPS, although CSP does significantly beat SeasonalNPTS on coverage ($p = 4 \times 10^{-6}$). The full battery of per-window paired Wilcoxon comparisons is reported in Appendix~\ref{app:sig}.

\noindent\textbf{Two claims hold cleanly across the audited suite.} First, both CSP variants are strictly ahead of all three NPTS-family methods (NPTS, SeasonalNPTS, DeepNPTS) on both mean CRPS rank and mean MQL rank --- true at every cell of the rank comparison in Table~\ref{tab:accuracy}. Second, both CSP variants significantly outperform DeepNPTS on every metric this paper rests on, with per-window paired Wilcoxon $p < 10^{-9}$ on CRPS, MQL, and coverage simultaneously.

CSP-Adaptive's mean within-dataset normalized CRPS is below DeepNPTS's, and CSP-Adaptive achieves mean $0.89$ coverage versus DeepNPTS's $0.66$. Full per-method absolute-score summary statistics (means and standard deviations of within-dataset CRPS ratios, normalized MQL, and coverage) are reported in Appendix~\ref{app:absscores}.

\subsection{Runtime}
Each method produces the same $380$ forecast records on this audit. CSP-Adaptive completes the full workload in $0.27$ wall-clock minutes; DeepNPTS takes $153.1$ minutes for the same workload, a $567\times$ slowdown. The five non-DeepNPTS methods together account for $1.6$ minutes; DeepNPTS alone is $99\%$ of total method-evaluation time. Figure~\ref{fig:pareto} places these numbers in an accuracy-runtime view: every training-free method sits in the upper-left zone (low CRPS rank, sub-minute runtime), and DeepNPTS is the only point in the upper-right --- worse on rank, and two and a half orders of magnitude slower. Full per-method wall-time numbers (sec/row, slowdown factors) are reported in Appendix~\ref{app:walltime}.

\begin{figure}[htbp]
\floatconts
  {fig:pareto}
  {\caption{Accuracy vs.\ runtime. Lower mean CRPS rank is better; the $x$-axis is log-scaled wall time. CSP-Adaptive and CSP-Fixed sit in the upper-left training-free zone; DeepNPTS is dominated on rank and isolated by two orders of magnitude on cost while also failing the coverage target.}}
  {\includegraphics[width=0.94\linewidth]{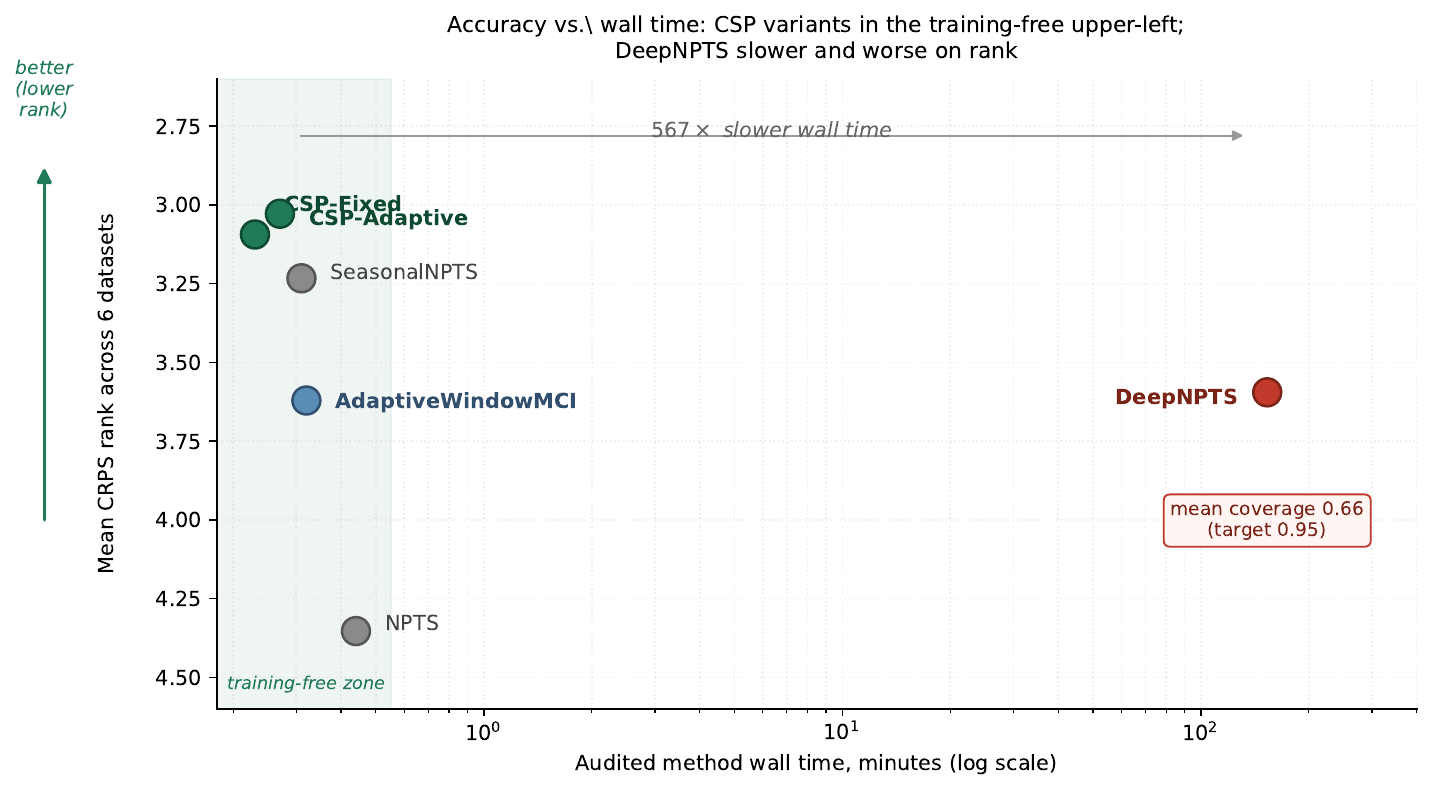}}
\end{figure}

\section{What Works, What Does Not, and Where}
\label{sec:diagnosis}
The CSP variants succeed because their two components correct different failure modes of empirical sampling: the signed-residual component recenters the predictive distribution around the latest seasonal level, while the seasonal pool keeps samples on realistic support. AdaptiveWindowMCI is a fast training-free fallback that uses no seasonal structure --- only a data-dependent recent window --- and serves as a sanity baseline against the seasonal-pool design choice. NPTS shows that pure recency-weighted empirical sampling is too coarse on hourly data (CRPS rank $4.35$); SeasonalNPTS, by contrast, is a strong baseline (CRPS rank $3.23$, coverage $0.91$) and is the closest non-CSP comparator on both metrics. DeepNPTS, despite being the most parameter-rich method in the comparator set, ends up worst on calibration: its mean coverage is $0.66$ against the nominal $0.95$ target, and its training cost is $567\times$ larger than CSP for the same forecast workload.

\section{Limitations and Scope}
\label{sec:limits}
\paragraph{Scope of the comparison set.}
The benchmark is scoped to non-parametric samplers: the NPTS family (NPTS, SeasonalNPTS, DeepNPTS) plus several training-free conformal and empirical samplers including CSP. We deliberately did not include classical distributional baselines (ETS, ARIMA, or TBATS with bootstrapped residuals), full-distribution conformal time-series methods built on parametric forecasters (e.g.\ EnbPI applied as a sampler), or other neural probabilistic forecasters such as DeepAR, Temporal Fusion Transformers, or N-HiTS. The goal of this paper is the head-to-head between training-free conformal sampling and DeepNPTS as the strongest published non-parametric neural sampler, and the choice of comparator set reflects that goal. The pattern of intervals systematically too narrow to attain nominal coverage that this paper documents for DeepNPTS is consistent with the broader prediction-interval failure documented across the M4 competition submissions \citep{grushkacockayne2020m4intervals}.

\paragraph{Benchmark size and selection.}
Each of the six datasets contributes $10$ series and $5$--$7$ rolling-origin windows per series, for $50$--$70$ paired forecast records per dataset and $380$ records per method overall. The series-count cap is set by the requirement that the DeepNPTS comparator be retrained from scratch on every series and remain auditable on a single workstation; full-series GluonTS configurations of the same datasets contain hundreds of series each and would not have been DeepNPTS-auditable in the available compute budget. The wall-clock comparison is a local CPU measurement; GPU hardware and different DeepNPTS training schedules will narrow the runtime gap, but two orders of magnitude is a wide enough margin that the qualitative conclusion is robust to those choices. The coverage gap is the more important comparison and does not depend on hardware.

\paragraph{Statistical and methodological caveats.}
The split-conformal residual pool used by CSP assumes exchangeability of calibration and test residuals, which is only approximately satisfied for non-stationary series; \citet{xu2021enbpi} and \citet{gibbs2021adaptive} provide drop-in extensions that we expect would further improve coverage on the most non-stationary datasets in the suite. The benchmark is restricted to the original DeepNPTS evaluation suite (electricity, exchange\_rate, solar\_energy, taxi, traffic, wikipedia), which gives a fair head-to-head against the comparator on its own design ground.

\paragraph{Method-design caveats.}
The CSP sampler treats forecast horizons independently within each predictive sample, so columns of the predictive sample matrix do not represent coherent forecast trajectories; downstream uses that require trajectory-level dependence (e.g.\ joint multi-horizon decision rules) would need a coupling step such as a block-bootstrap variant. The signed-residual pool is uniform over the calibration window rather than studentized or recency-weighted; on heteroskedastic or strongly drifting series, scale normalisation by a rolling seasonal MAD or variance is a natural extension we did not test here. We also did not run sensitivity analyses for the sample budget $B=100$, the recency rate $\lambda$, or the calibration fraction $\rho$; these are fixed across all datasets in the audited run.

\section{Conclusion}
\label{sec:conclusion}
On the six datasets where DeepNPTS was originally benchmarked, a simple training-free conformal sampler significantly outperforms it on CRPS ($p \approx 4 \times 10^{-10}$ per-window paired Wilcoxon), on normalized mean quantile loss ($p \approx 7 \times 10^{-10}$), and on empirical $95\%$ coverage ($p \approx 8 \times 10^{-45}$, mean $0.89$ vs $0.66$), while running $567\times$ faster on CPU. The coverage gap is the most decision-critical: under the calibration-then-sharpness reporting protocol of \citet{gneiting2014probabilistic}, a method whose nominal $95\%$ interval misses the truth roughly a third of the time fails the calibration prerequisite and is inadmissible for any setting that takes the nominal level seriously.

We frame this as evidence that training-free seasonal residual samplers should be included as mandatory baselines when evaluating learned non-parametric forecasters, and that future probabilistic-forecasting benchmarks should report empirical coverage at the nominal level alongside any sharpness score. The result is specific to the audited DeepNPTS configuration on this benchmark and is not a general claim about neural probabilistic forecasting.

\acks{The authors thank the maintainers of the GluonTS toolkit for the public datasets and the reference DeepNPTS implementation used in this benchmark.}

\appendix

\section{Implementation Details}
\label{app:impl}
This appendix documents the parameter choices made for every method in the audited suite.

\paragraph{CSP-Fixed and CSP-Adaptive.}
Algorithm~\ref{alg:csp} states the construction. The CSP-Adaptive weight rule ($w_h \in \{0, 0.3, 0.5\}$) was chosen ahead of seeing test scores and is fixed across all datasets; we did not tune it per dataset. Hyperparameters: $\rho = 0.5$ (calibration fraction), $\lambda = 0.01$ (exponential recency rate, in normalized time-index units), $B = 100$ predictive samples per forecast record (matched across all training-free methods).

\paragraph{Rolling-origin protocol.}
For each dataset the audit runs $5$--$7$ rolling-origin windows with stride equal to the forecast horizon $H$ (non-overlapping forecast targets). Horizons: $H=24$ for hourly datasets, $H=30$ for daily datasets. The same window indices are used across all methods so per-dataset comparisons are paired.

\paragraph{DeepNPTS.}
Run via \texttt{gluonts.torch.model.deep\_npts.DeepNPTSEstimator} with $50$ epochs, batch size $32$, two hidden layers of $40$ units, learning rate $10^{-4}$, $100$ batches/epoch, context multiplier $4$ (capped at $512$ steps), $200$ inference samples. No early stopping, no per-dataset tuning. CPU only.

\paragraph{Other methods.}
NPTS and SeasonalNPTS: GluonTS reference implementations, default parameters. AdaptiveWindowMCI: ACF-based seasonality detector (threshold $0.3$) selecting a data-dependent recent window. All non-DeepNPTS methods use experiment-driver seed $0$ and internal RNG seed $42$.

\paragraph{Per-dataset preprocessing.}
No normalization, gap-filling, or frequency conversion. Series shorter than $H+10$ observations are filtered out.

\section{Dataset Suite}
\label{app:datasets}
\begin{table}[htbp]
\floatconts
  {tab:datasets}
  {\caption{Audited dataset suite. Records are forecast records per method.}}
  {\begingroup
   \small
   \setlength{\tabcolsep}{6pt}
   \renewcommand{\arraystretch}{0.95}
   \begin{tabular}{lrr}
\toprule
Dataset & Records & Horizon \\
\midrule
electricity & 70 & 24 \\
exchange\_rate & 50 & 30 \\
solar\_energy & 70 & 24 \\
taxi & 70 & 24 \\
traffic & 70 & 24 \\
wikipedia & 50 & 30 \\
\bottomrule
\end{tabular}

   \endgroup}
\end{table}

\section{Per-Method Absolute Scores}
\label{app:absscores}
Table~\ref{tab:absolute} reports the per-method mean and standard deviation of within-(dataset, series, window) normalized CRPS, normalized MQL, and empirical coverage at the $0.95$ nominal level.

\begin{table}[htbp]
\floatconts
  {tab:absolute}
  {\caption{Absolute-score summary across the $6$ datasets. CRPS is reported as a within-dataset ratio to the cross-method median to control for the wide range of dataset scales. MQL is the normalized mean quantile loss. Coverage targets $0.95$.}}
  {\begin{tabular}{lrrrrr}
\toprule
 & \multicolumn{2}{c}{\textbf{CRPS (rel.\ to row median)}} & \multicolumn{2}{c}{\textbf{Normalized MQL}} & \textbf{Coverage} \\
\cmidrule(lr){2-3}\cmidrule(lr){4-5}\cmidrule(lr){6-6}
Method & Mean & Std & Mean & Std & Mean \\
\midrule
\textbf{CSP-Adaptive} & 0.915 & 0.062 & 0.405 & 0.472 & 0.888 \\
\textbf{CSP-Fixed} & 0.915 & 0.062 & 0.405 & 0.471 & 0.891 \\
AdaptiveWindowMCI & 1.044 & 0.090 & 0.528 & 0.600 & 0.695 \\
DeepNPTS & 1.257 & 0.393 & 0.388 & 0.358 & 0.662 \\
SeasonalNPTS & 0.963 & 0.150 & 0.442 & 0.551 & 0.908 \\
NPTS & 1.432 & 0.507 & 0.438 & 0.407 & 0.950 \\
\bottomrule
\end{tabular}
}
\end{table}

\section{Per-Window Paired Significance Tests}
\label{app:sig}
Table~\ref{tab:sig} reports the full battery of per-window paired Wilcoxon signed-rank tests on within-(dataset, series, window) normalized CRPS and MQL (each method's score divided by the cross-method median for that window) and on coverage. The headline numbers are reported in Section~\ref{sec:sig}; this appendix records the comparison set in full.

\begin{table}[htbp]
\floatconts
  {tab:sig}
  {\caption{Per-window paired Wilcoxon signed-rank tests ($n \approx 380$ paired records per comparison). Bold $p$-values reach $p<0.05$. CRPS and MQL are normalized within each (dataset, series, window) by the median across methods.}}
  {\begin{tabular}{lr}
\toprule
Comparison & per-window paired Wilcoxon $p$ \\
\midrule
\multicolumn{2}{l}{\textit{CRPS, normalized within (dataset, series, window) by median across methods, $A < B$}} \\
CSP-Adaptive $<$ DeepNPTS & $\mathbf{4.4e-10}$ \\
CSP-Fixed $<$ DeepNPTS & $\mathbf{2.5e-10}$ \\
AdaptiveWindowMCI $<$ DeepNPTS & $0.1539$ \\
CSP-Adaptive $<$ SeasonalNPTS & $0.1702$ \\
CSP-Adaptive $<$ NPTS & $\mathbf{1.3e-28}$ \\
\midrule
\multicolumn{2}{l}{\textit{Normalized mean quantile loss, $A < B$}} \\
CSP-Adaptive $<$ DeepNPTS & $\mathbf{6.6e-10}$ \\
CSP-Fixed $<$ DeepNPTS & $\mathbf{6.1e-10}$ \\
AdaptiveWindowMCI $<$ DeepNPTS & $0.1789$ \\
\midrule
\multicolumn{2}{l}{\textit{Coverage (closer to nominal $0.95$), $A > B$}} \\
CSP-Adaptive $>$ DeepNPTS & $\mathbf{8.0e-45}$ \\
CSP-Fixed $>$ DeepNPTS & $\mathbf{1.1e-44}$ \\
AdaptiveWindowMCI $>$ DeepNPTS & $0.8570$ \\
CSP-Adaptive $>$ SeasonalNPTS & $\mathbf{3.9e-06}$ \\
CSP-Adaptive $>$ NPTS & $0.7237$ \\
\bottomrule
\end{tabular}
}
\end{table}

\section{Per-Method Wall Time}
\label{app:walltime}

\begin{table}[htbp]
\floatconts
  {tab:walltime}
  {\caption{Audited method wall time. Dataset download/loading and final summary writing are excluded.}}
  {\begin{tabular}{lrrrrr}
\toprule
Method & Wall min. & Rows & Sec./row & Slowdown vs fastest & Datasets \\
\midrule
CSP-Fixed & 0.23 & 380 & 0.019 & 1.0$\times$ & 6 \\
CSP-Adaptive & 0.27 & 380 & 0.022 & 1.2$\times$ & 6 \\
SeasonalNPTS & 0.31 & 380 & 0.025 & 1.3$\times$ & 6 \\
AdaptiveWindowMCI & 0.32 & 380 & 0.026 & 1.4$\times$ & 6 \\
NPTS & 0.44 & 380 & 0.036 & 1.9$\times$ & 6 \\
DeepNPTS & 153.10 & 380 & 12.500 & 665.7$\times$ & 6 \\
\bottomrule
\end{tabular}
}
\end{table}

\clearpage
\begingroup
\small
\setlength{\bibsep}{2pt}
\bibliographystyle{plainnat}
\bibliography{cp_ts_benchmark_refs}
\endgroup

\end{document}